\pdfoutput=1

\documentclass[11pt]{article}

\usepackage[preprint]{acl}

\usepackage{times}
\usepackage{latexsym}

\usepackage[T1]{fontenc}

\usepackage[utf8]{inputenc}

\usepackage{microtype}

\usepackage{inconsolata}

\usepackage{booktabs}
\usepackage{multirow}
\usepackage{url}
\usepackage{hyperref}
\usepackage{graphicx}
\usepackage{amsmath}
\usepackage[originalparameters]{ragged2e}
\usepackage{subfigure}
\usepackage{authblk}
\usepackage[skip=3pt]{caption}
\title{``Is There Anything Else?'': Examining Administrator Influence on Linguistic Features from the Cookie Theft Picture Description Cognitive Test}

\author[1]{Changye Li}
\author[2]{Zhecheng Sheng}
\author[1]{Trevor Cohen}
\author[2]{Serguei Pakhomov}

\affil[1]{University of Washington}
\affil[2]{University of Minnesota}

\affil[1]{\texttt{\{changyel,cohenta}\}@uw.edu}
\affil[2]{\texttt{\{sheng136, pakh0002}\}@umn.edu}

\begin{document}
\maketitle
\begin{abstract}

Alzheimer's Disease (AD) dementia is a progressive neurodegenerative disease that negatively impacts patients' cognitive ability. Previous studies have demonstrated that changes in naturalistic language samples can be useful for early screening of AD dementia. However, the nature of language deficits often requires test administrators to use various speech elicitation techniques during spontaneous language assessments to obtain enough propositional utterances from dementia patients. This could lead to the ``observer's effect'' on the downstream analysis that has not been fully investigated. Our study seeks to quantify the influence of test administrators on linguistic features in dementia assessment with two English corpora the ``Cookie Theft'' picture description datasets collected at different locations and test administrators show different levels of administrator involvement. Our results show that the level of test administrator involvement significantly impacts observed linguistic features in patient speech. These results suggest that many of significant linguistic features in the downstream classification task may be partially attributable to differences in the test administration practices rather than solely to participants' cognitive status. The variations in test administrator behavior can lead to systematic biases in linguistic data, potentially confounding research outcomes and clinical assessments. Our study suggests that there is a need for a more standardized test administration protocol in the development of responsible clinical speech analytics frameworks.\footnote{Our code is available at \url{https://github.com/LinguisticAnomalies/turns}}
\end{abstract}

\section{Introduction}

Alzheimer's Disease (AD) dementia is a neurodegenerative disease that causes progressive decline in cognitive function. Even though AD currently has no cure, a timely diagnosis is imperative to alleviate negative consequences of delayed or absent diagnosis including emergency events, family strife, and exposure to scam artists praying on the vulnerable \citep{doi.org/10.1111/psyg.12095}. Changes in naturalistic language samples collected from individuals at high-risk for dementia have been identified as one of the early signs of AD \citep{almor1999alzheimer, blanken1987spontaneous,bucks2000analysis}, showing its potential as an early screening tool. However, analyzing speech samples is labor-intensive and time-consuming. Contemporary studies predominately focus on automated prediction and detection of such changes with language models with considerable success in distinguishing the speech of dementia patients and healthy controls (for recent reviews, see \citet{SHI2023115538, ding2024speech}). Despite these advances, this line of research often faces the limited data availability. As noted in \citet{SHI2023115538}, the majority of prior work focuses on analyzing naturalistic speech samples using the transcripts of ``Cookie Theft'' picture description cognitive task produced by English-speaking cohorts in the Pitt corpus \citep{10.1001/archneur.1994.00540180063015}.

While several prior studies have focused on connected speech from non-English speaking participants (e.g., French \citep{ROUSSEAUX20103884}, Spanish \citep{10.3389/fnagi.2020.00270}, and German \citep{weiner16_interspeech}), a very limited discussion has been held in prior literature on the influence of test administrators. Similarly, methods for data collection, such as optimal sample duration, distance to the microphone, and presence of background noise, have not been standardized \citep{doi:10.1080/02687038.2012.654933}. In addition, the impaired communication ability of people with dementia \citep{doi:10.1212/01.wnl.0000210435.72614.38, hier1985language, rousseaux2010analysis} creates additional barriers for their caregivers \citep{Eggenberger2012CommunicationST, banovic2018communication}. This could also extend to neuropsychological assessment batteries such as picture description tasks, which are used extensively by speech-language pathologists in the management of clients with language disorders, including aphasia and dementia \citep{cummings2019describing, berube2019stealing}. Prior works have demonstrated that test administrators often perform a variety of speech elicitation techniques to extract additional propositions from aphasic patients \citep{menn1989cross, CAPLAN1998184}. As a number of studies have argued in favor of a similarity of linguistic behavior in patients with dementia and aphasia \citep{gewirth1984altered,Nicholas1985EmptySI,blanken1987spontaneous,Gumus2024LinguisticCI}, similar elicitation strategies may be employed when collecting speech samples from dementia patients. This could lead to the ``observer effect'' \citep{labov1973sociolinguistic} in feature values as many distinct linguistic features are sensitive to the length of the text sample. A previous study \citep{10.1159/000533423} demonstrated that sample length is important for extracting the various language features of AD by analyzing the speech samples (e.g., public interviews, talk shows and public speeches) from cognitively healthy public figures and those diagnosed with AD dementia. However, this previous study did not address the influence of interviewers and their speech elicitation techniques on collected speech. The impact of test administrators/interviewers and the resulting reliability of linguistic features in clinical settings also remains understudied. This less-discussed gap is particularly concerning given the potential for these factors to introduce systematic biases in the assessment of cognitive decline.

To address this limitation, our study seeks to quantify the influence of test administrators on speech collected with the ``Cookie Theft'' picture description task. Specifically, we analyze the quantity and distribution of part-of-speech (POS) tags in task transcripts collected from participants residing in two distinct United States locations: Pennsylvania and Wisconsin. We anticipate that test administrators employ significantly more interactions to elicit speech from dementia patients compared to healthy controls, which may contribute to patients with dementia producing linguistic patterns found to be associated with dementia, such as increased use of repetitions \citep{hier1985language}, higher pronoun usage \citep{almor1999alzheimer}, and elevated lexical frequency \citep{bucks2000analysis} when compared to healthy controls. We analyze the Pitt corpus and the Wisconsin Longitudinal Study (WLS) \citep{10.1093/ije/dys194} datasets from the Dementia Bank. Both employ the ``Cookie Theft'' picture description task from the Boston Diagnositc Aphasia Examination \citep{goodglass1983boston}. We aim to quantify the extent to which the linguistic features commonly attributed to dementia patients may be artifacts of the data collection and test administration process.

The contributions of this work can be summarized as follows: a) we examine patterns in how test administrator involvement may relate to linguistic features observed in patient speech and their association with dementia vs. control classification; and b) our analyses raise questions about how variations in test administrator behavior might interact with linguistic patterns in clinical assessments. These observations point to opportunities for future research to investigate the role of test administration in linguistic analyses and clinical assessments.

\section{Related Work}

Verbal production tasks are common neuropsychological assessments for measuring language and executive retrieval functions, with the category fluency task being one of the most widely utilized in clinical settings. In this task, participants are asked to generate exemplars of specific semantic categories -- such as animals or food -- in a given time. While the category fluency task has demonstrated the diagnostic utility for AD screening \citep{monsch1992comparisons, cerhan2002diagnostic}, these assessments are typically conducted in controlled clinical settings and often require longitudinal observation before a final diagnosis can be made. Such controlled testing environments can be insensitive to naturalistic language patterns \citep{sabat1994language} and may miss early signs of linguistic deficits that manifest in daily communications \citep{crockford1994assessing}. In contrast, spontaneous speech has proven to be a valuable source of information for assessing an individual's cognitive state \citep{bucks2000analysis}.

\begin{figure}
    \centering
    \includegraphics[width=0.9\linewidth]{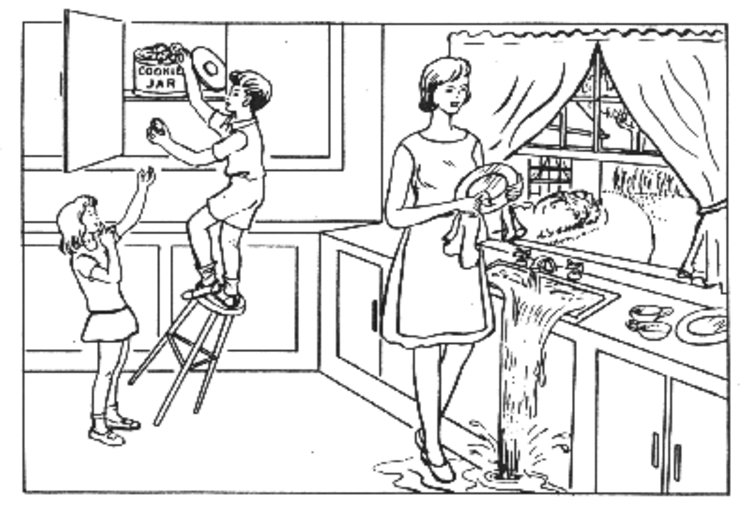}
    \caption{The ``Cookie Theft'' picture description stimuli.}
    \label{fig:cookie}
\end{figure}

The ``Cookie Theft'' picture description task (Figure~\ref{fig:cookie}) is designed to elicit speech samples in pathological cohorts. Participants are asked to describe everything they observe in a picture where two children collaborate to secretly take cookies from a high cupboard shelf, while their mother is preoccupied washing dishes. Previous studies using \textit{statistical} analyses have demonstrated many linguistic anomalies associated with AD progression, such as increased use of repetitions \citep{hier1985language}, higher pronoun usage \citep{almor1999alzheimer}, and elevated lexical frequency \citep{bucks2000analysis, cummings2019describing} compared to healthy controls. Supervised machine/deep learning methods, including transformer-based \citep{NIPS2017_3f5ee243} neural language models can learn to distinguish subtle linguistic characteristics between dementia patients and healthy controls with impressive classification performance (for a review , see \citet{ding2024speech}). However, such models bring an additional challenge -- often the best-performing models (i.e., neural language models) are least transparent, and the less-accurate models (i.e., statistical models) are easier to explain.  Limited interpretability could obscure the bias, which is particularly concerning in clinical artificial intelligence development \citep{reddy2022explainability}.

Building upon the previous findings that longer speech is important to extract distinguishable linguistic features \citep{10.1159/000533423} and interaction patterns between speakers are predictive of the downstream classification task \citep{farzana-parde-2022-interaction}, we build statistical models to investigate the role of test administrator behavior in the manifestation of linguistic markers associated with dementia. We show that the level of test administrator's engagement significantly impacts the linguistic features observed in the patients' speech.

\section{Method}

\subsection{Data}

\begin{table*}[htbp]
\centering
\small
\begin{tabular}{@{}|cl|cc|cc|@{}}
\toprule
\multicolumn{2}{|c|}{\multirow{2}{*}{\textbf{Characteristics}}}    & \multicolumn{2}{c|}{\textbf{Pitt}}    & \multicolumn{2}{c|}{\textbf{WLS}}    \\ \cmidrule(l){3-6} 
\multicolumn{2}{|c|}{} & \multicolumn{1}{c|}{Control} & Dementia & \multicolumn{1}{c|}{Control} & Dementia \\ \midrule
\multicolumn{1}{|c|}{\multirow{2}{*}{Gender (\%)}} & Female & \multicolumn{1}{c|}{57 (59.4)} & 99 (68.3) & \multicolumn{1}{c|}{523 (51.4)} & 63 (41.4) \\ \cmidrule(l){2-6} 
\multicolumn{1}{|c|}{}  & Male & \multicolumn{1}{c|}{39 (40.6)} & 46 (31.7) & \multicolumn{1}{c|}{494 (48.6)} & 89 (58.6) \\ \midrule
\multicolumn{2}{|c|}{\# of transcripts} & \multicolumn{1}{c|}{182} & 214 & \multicolumn{1}{c|}{1017} & 152 \\ \midrule
\multicolumn{2}{|c|}{Age (mean (SD))} & \multicolumn{1}{c|}{64.1 (7.9)} & 71.5 (8.63) & \multicolumn{1}{c|}{70.30 (4.14)} & 70.20 (5.75) \\ \midrule
\multicolumn{2}{|c|}{Education (mean (SD))} & \multicolumn{1}{c|}{13.9 (2.4)} & 12.3 (2.8) & \multicolumn{1}{c|}{13.77 (3.01)} & 12.64 (2.16) \\\bottomrule
\end{tabular}
\caption{Basic characteristics of the Pitt corpus and the WLS corpus before propensity score matching.}
\label{tab:data}
\end{table*}

We use two publicly available datasets resulting from deploying the ``Cookie Theft'' picture description task during data collection: a) the Pitt corpus\footnote{\url{https://dementia.talkbank.org/access/English/Pitt.html}} and b) the WLS\footnote{\url{https://dementia.talkbank.org/access/English/WLS.html}} corpus. The Pitt corpus includes recordings and corresponding transcripts from 319 participants. 102 out of 319 participants were classified as control subjects and 204 participants as patients categorized with any AD-related label. Specifically, we restricted the original Pitt corpus to a subset of 169 patients with an assignment of probable AD dementia and 99 healthy controls, resulting in 214 and 182 transcripts for AD patients and healthy controls, respectively. 

The WLS is a large-scale, extended longitudinal study of a random sample of 10,317 men and women who graduated from Wisconsin high schools in 1957. The WLS participants were interviewed up to 6 times between 1957 and 2011. Several nueropsychological tests, including letter fluency task and category fluency task were administered in both 2004 and 2011. The ``Cookie Theft'' picture description task was introduced in 2011. While the WLS participants were interviewed with Telephone Interview for Cognitive Status-modified (TICS-m) for a clinical proxy diagnosis in 2020, we decide to follow a prior study \citep{guo2021crossing} to build a ``noisy'' label with statistically determined age- and education-adjusted thresholds of 16, 14, and 12 for participants in $<$ 60, 60-79, and $>$ 79 age ranges for the category fluency score, respectively. This addresses a critical temporal aspect in AD assessment, particularly given the 9-year gap between speech data collection and clinical assessment in the WLS dataset, contrasting with the Pitt corpus where participants were diagnosed at the time of speech collection. In supporting this approach, the category fluency task, administered concurrently with the ``Cookie Theft'' picture description task in the WLS corpus, has demonstrated the diagnostic utility on discriminating AD patients and healthy controls, with sensitivity of 0.88 and specificity of 0.96 \citep{canning2004diagnostic}. Additionally, the number of WLS participants who completed both the cognitive tests and follow-up clinical interview remained particularly small ($<$ 35 labeled dementia patients), potentially limiting the statistical power of our study.

As a result, we restrict the original WLS dataset to a total of 1,169 participants (1,017 healthy controls and 152 dementia cased patients) who a) agreed to participant in the ``Cookie Theft'' picture description task and category fluency test in 2011; b) had not been diagnosed with a mental illness at the time of interview; and c) did not previously have a stroke at the time of the interview. Given the fact that the Pitt corpus contains dementia labels obtained from clinical assessments conducted concurrently with the picture description task, we consider this to be an example of dementia \textit{detection}. In contrast, the WLS dataset represents the case of dementia \textit{prediction}. Data characteristics are provided in Table~\ref{tab:data}.


\subsection{Preprocessing}

We perform transcript pre-processing using TRESTLE (\textbf{T}oolkit for \textbf{R}eproducible \textbf{E}xecution of \textbf{S}peech \textbf{T}ext and \textbf{L}anguage \textbf{E}xperiments) \citep{li2023trestle} for both participants and test administrators. Specifically, we remove non-ASCII characters, unintelligible words, and non-speech artifacts event descriptions or gestures. We also retain the utterances from participants in a relatively ``raw'' state,  in which we preserve repetitions, invited interruptions, and speech repairs (self-revisions).

\subsection{Topics Analysis}

We segment the utterances from test administrators into individual sentences and remove the duplicates to establish a clean dataset for analysis. These utterances are then clustered based on frequency in each diagnostic group to understand the predominant conversation topics.

\subsection{Linguistic feature extraction}

Following the established evidence \citep{bucks2000analysis, almor1999alzheimer, hier1985language, cummings2019describing, blanken1987spontaneous}, we focus our the analysis of part-of-speech (POS) tags, lexical frequency (LF), and type-to-token ratio (TTR) on utterances from participants in the Pitt and WLS corpora. We extract the counts of each POS tag for each transcript using spaCy\footnote{https://spacy.io/} with RoBERTa \citep{Liu2019RoBERTaAR} as the base model\footnote{See Table~\ref{tab:pos} in Appendix for the full list of POS tags analyzed in this study.}. The log LF of each transcript is calculated using the SUBTLEX$_{\text{us}}$ corpus \citep{brysbaert2009moving}. Tokens that do not appear in the SUBTLEX$_{\text{us}}$ corpus are removed as out-of-vocabulary items. TTR quantifies lexical diversity in speech samples, calculated as the proportion of unique words to total words in the transcript. We also count the number of clauses in each transcript. In this study, we define a clause as a syntactic unit centered around a verb that expresses a proposition. As a proxy of syntactic complexity \citep{CAPLAN1998184}, clause count has been shown to be a sensitive linguistic feature for detecting dementia from spoken samples \citep{doi:10.1080/02687038.2012.654933, pakhomov2011computerized}.

Additionally, we define \textit{turn} as the number of utterances from either participants (denoted as \texttt{par}\_\texttt{turns}) or test administrators (denoted as \texttt{inv}\_\texttt{turns}) in each transcript. We extract the number of turns from test administrators from transcripts for follow-up propensity score matching (PSM).

\subsection{Propensity score matching}

Propensity score matching (PSM) \citep{doi:10.1080/00273171.2011.568786} is a statistical matching method to estimate the effect of a treatment by accounting for the covariates that predict receiving the treatment. PSM assigns a propensity score, which is the probability of treatment assignment conditional on the observed covariates. This conditional probability, serving as a balancing score, matches each individual in the treatment group to an individual in the control group in controlled experiments. 

\citet{luz20_interspeech} introduces the AD Recognition through Spontaneous Speech (ADReSS) Challenge, providing researchers with the first available benchmark that is acoustically pre-processed and balanced in terms of age and gender, both of which are risk factors for AD \citep{RUITENBERG2001575, van2005epidemiology}. However, it does not take into account the following possible confounding factors: a) education level, (lower education level is a risk factor of dementia later in life and contributes to the lower linguistic ability) \citep{snowdon1996linguistic, ngandu2007education, nguyen2016instrumental, caamano2006education}; and b) the influence of test administrators, who may perform a variety of speech elicitation techniques to extract enough propositions from patients \citep{menn1989cross, CAPLAN1998184} in a constrained task, such as the ``Cookie Theft'' picture description task.

To address these concerns, we match the Pitt and the WLS corpora on: a) years of education received, and b) the number of turns from test administrators using PSM. This resulted in a balanced Pitt corpus with 167 transcripts for both dementia patients and healthy controls, and a balanced WLS corpus containing 152 transcripts for both dementia patients and healthy controls.

\subsection{Statistical models}

We apply z-score normalization on the POS tags, lexical frequency and TTR extracted from each transcript and treat the number of turns from test administrators as the random effects. We split the original and the matched Pitt corpus into 70/30 training/test split. We fit a generalized linear mixed models on the \textit{matched} Pitt training split where we treat the number of turns from test administrators (\texttt{inv}\_\texttt{turns}) as random effects. Our preliminary results show that fitting such a model for \textit{matched} WLS data results in singularity (i.e., the random effects of \texttt{inv}\_\texttt{turns} variance-covariance matrix is of \textit{less than full rank}). Therefore we decide to fit generalized linear model on the WLS corpus. In addition, we compare the interaction model (models with interaction terms between \texttt{inv}\_\texttt{turns} and each linguistic feature) and na\"ive models (models without interaction terms) and apply backward selection using  Akaike’s Information Criteria (AIC) \citep{akaike1998information}. AIC is an information-theoretic approach that estimates the distance between candidate models and the true model on a log-scale, which selects a parsimonious approximating model for the observed data. Our preliminary results show that interaction models achieve better fit with lower AIC. We then continue our analysis with the resulting interaction model for Pitt corpus ($\mathcal{M}_{\text{pitt}}$) and WLS corpus ($\mathcal{M}_{\text{wls}}$).


We also perform cross validation on each dataset to test for internal validity. Specifically, we assess the classification performance of $\mathcal{M}_{\text{pitt}}$ on both the matched Pitt test split and the matched WLS corpus and $\mathcal{M}_{\text{wls}}$ on the matched Pitt test split, respectively.

\section{Results}

The results of PSM for the Pitt and the WLS corpus can be found in Table~\ref{tab:total_pitt} and Table~\ref{tab:total_wls} in Appendix, respectively. We observed that many linguistic features preserved imbalance even after PSM, with standardized mean difference (SMD) $>0.1$ \citep{zhang2019balance}. It should be noted that SMD does not indicate the differences in the direction of the scale \citep{chandler2019cochrane} (i.e., cannot substitute the p-value from significance testing). We also observed that the WLS participants obtained a higher level of education than the Pitt participants (one-sided Wilcoxon rank sum test p-value $<$ 0.001). These observations suggest that additional, potentially unaccounted-for variability may be influencing the results. Thus we proceeded with further quantitative and qualitative analyses.

\subsection{Topics analysis}

We found that test administrators' utterances usually cover the following topics: a) initiation of the task (e.g., ``and there's a picture'' and ``what's going on in this picture''); b) acknowledgment of progress (e.g., ``okay''); c) speech elicitation (e.g., ``anything else'', ``if you see anything else'' and ``is there anything else''); and d) ending the task (e.g., ``alright'', ``thank you'', ``that's fine" and "good''). For the Pitt corpus, test administrators said ``anything else?'' more frequently to dementia patients (18 times) than to healthy controls (10 times). In contrast, the WLS test administrators used the same level of speech elicitation for both groups (dementia patients: 2 times; healthy controls: 2 times). 


\subsection{Test administrator interaction styles}

\begin{table}[htbp]
\centering
\resizebox{\columnwidth}{!}{%
\begin{tabular}{@{}ccc|p{2cm}|p{2cm}@{}}
\toprule
\multicolumn{3}{c|}{\textbf{Dataset/Condition}}  & \textbf{\RaggedRight Participants' turns (mean (SD))} & \textbf{\RaggedRight Test administrators' turns (mean (SD))} \\ \midrule
\multicolumn{1}{c|}{\multirow{4}{*}{Pitt}} & \multicolumn{1}{c|}{\multirow{2}{*}{Before}} & Control & 13.55 (6.04) & 3.16 (1.77) \\ \cmidrule(l){3-5} 
\multicolumn{1}{c|}{}  & \multicolumn{1}{c|}{matching}                  & Dementia  & 13.54 (6.98) & 6.10 (4.48) \\ \cmidrule(l){2-5} 
\multicolumn{1}{c|}{}  & \multicolumn{1}{c|}{\multirow{2}{*}{After}} & Control & 13.44 (5.97) & 3.34 (1.73) \\ \cmidrule(l){3-5} 
\multicolumn{1}{c|}{}  & \multicolumn{1}{c|}{matching}                  & Dementia &  12.38 (5.60)&  4.38 (1.85) \\ \midrule
\multicolumn{1}{c|}{\multirow{4}{*}{WLS}} & \multicolumn{1}{c|}{\multirow{2}{*}{Before}} & Control & 14.39 (7.91) &  0.75 (1.53)\\ \cmidrule(l){3-5} 
\multicolumn{1}{c|}{}  & \multicolumn{1}{c|}{matching}                  & Dementia & 11.97 (7.04) & 0.82 (1.79) \\ \cmidrule(l){2-5} 
\multicolumn{1}{c|}{}   & \multicolumn{1}{c|}{\multirow{2}{*}{After}} & Control & 13.80 (7.76) & 0.82 (1.62) \\ \cmidrule(l){3-5} 
\multicolumn{1}{c|}{}  & \multicolumn{1}{c|}{match} & Dementia & 11.97 (7.04) & 0.82 (1.79) \\ \bottomrule
\end{tabular}
}
\caption{The number of turns from participants and test administrators in the Pitt and the WLS corpus, before and after matching.}
\label{tab:turn}
\end{table}

We observed a moderate negative correlation (Spearman's $\rho=-0.481$) between the number of turns used by Pitt test administrators and participants' Mini-Mental State Examination (MMSE) scores. Pitt test administrators interacted more with dementia patients who had lower MMSE scores, likely in an effort to elicit sufficient speech for analysis. As shown in Table~\ref{tab:turn}, Pitt test administrators used 3 more turns on dementia patients compared to healthy controls whereas the WLS test administrators uses similar number of turns on both diagnostic groups.

\subsection{Quantifying the administrator effect}

\begin{figure*}[htbp]
\centering
\subfigure[The coefficients of $\mathcal{M}_{\text{pitt}}$ after backward selection with AIC]{
  \includegraphics[width=0.48\textwidth]{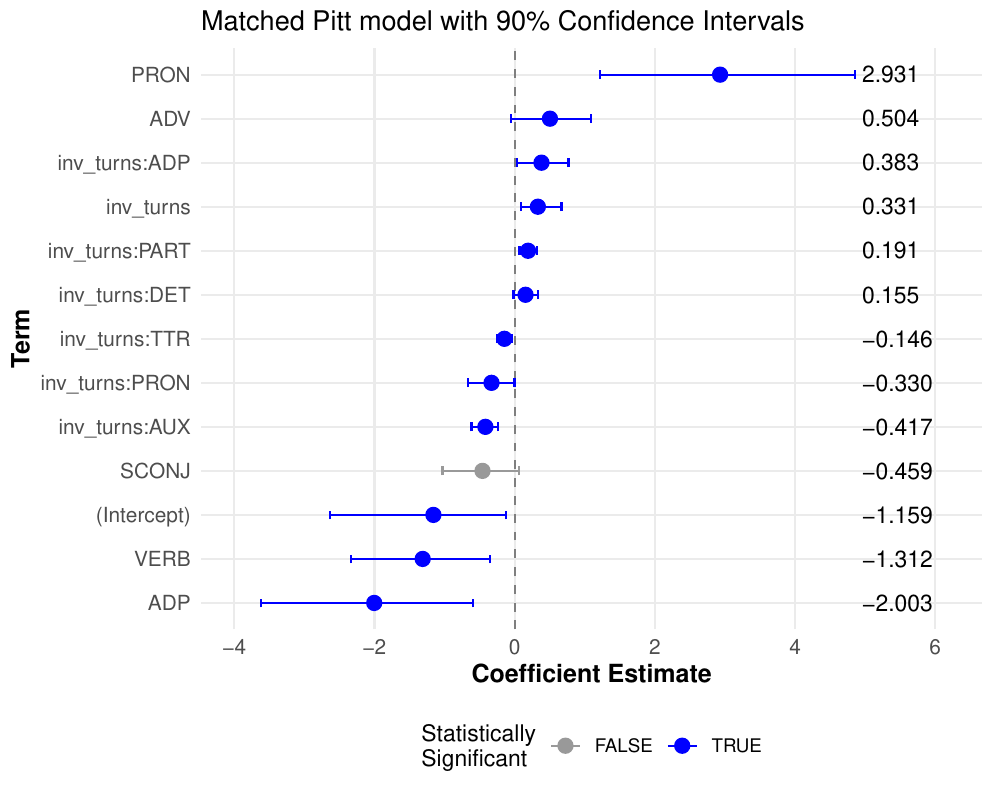}
  \label{fig:pitt}
}
\hfill
\subfigure[The coefficients of $\mathcal{M}_{\text{wls}}$ after backward selection with AIC]{
  \includegraphics[width=0.48\textwidth]{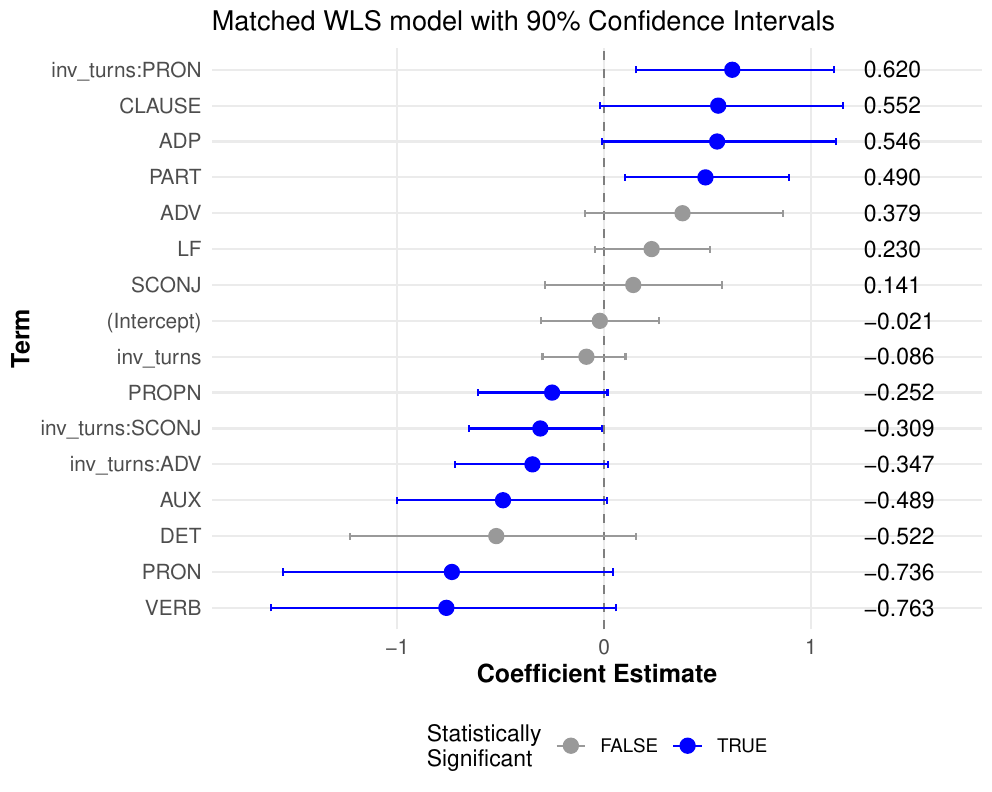}
  \label{fig:wls}
}
\caption{The estimated coefficients and the corresponding 90\% confidence intervals of $\mathcal{M}_{\text{pitt}}$ and $\mathcal{M}_{\text{wls}}$. The blue points and ranges indicate that the confidence interval does not cross zero, suggesting the estimate is statistically significant, whereas the dark gray points and ranges indicate that the confidence interval crosses zero, suggesting the estimate is not statistically significant.}
\label{fig:coef}
\end{figure*}

\paragraph{The Pitt model}As shown in Figure~\ref{fig:pitt}, we found that the number of test administrators' turns remain positive and significant ($\beta=0.331$, p-value $<$ 0.05) in the $\mathcal{M}_{\text{pitt}}$, suggesting that a more interactive test administrator dynamic is associated with a higher probability of developing dementia. We also observed that pronoun usage ($\beta=2.93$, p-value $<$ 0.001) showed a strong positive association with a higher probability of developing dementia. Interestingly, we observed significant interactions between test administrators' turns and various linguistic features, including TTR ($\beta=-0.146$, p-value $<$ 0.001), the usage of pronoun usage ($\beta=-0.330$, p-value $<$ 0.05), auxiliary ($\beta=-0.417$, p-value $<$ 0.001), adposition ($\beta=0.382$, p-value $<$ 0.05), and particle ($\beta=0.191$, p-value $<$ 0.001). 

\paragraph{The WLS model} As showed in Figure~\ref{fig:wls}, we observed fewer significant predictors in $\mathcal{M}_{\text{wls}}$. Interestingly, we observed that, while the usage of pronoun ($\beta=-0.76$, p-value $<$0.1) showed significantly negative association with having a dementia diagnosis, its interactions terms with the number of test administrators' turns demonstrated an \textit{opposite} directional effects ($\beta=0.620$, p-value $<$ 0.05).

\begin{figure*}[htbp]
\centering
\subfigure[The effect of the interaction term between pronoun usage and \texttt{inv}\_\texttt{turns}]{
  \includegraphics[width=0.48\textwidth]{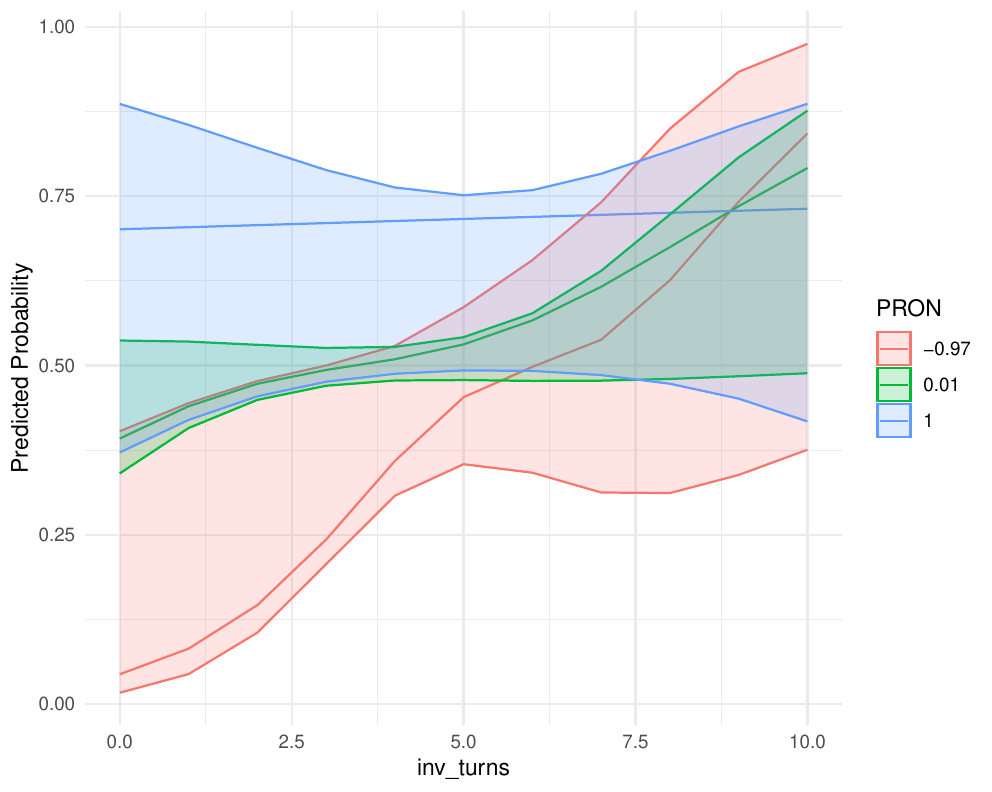}
  \label{fig:pron}
}
\hfill
\subfigure[The effect of the interaction term between TTR and \texttt{inv}\_\texttt{turns}]{
  \includegraphics[width=0.48\textwidth]{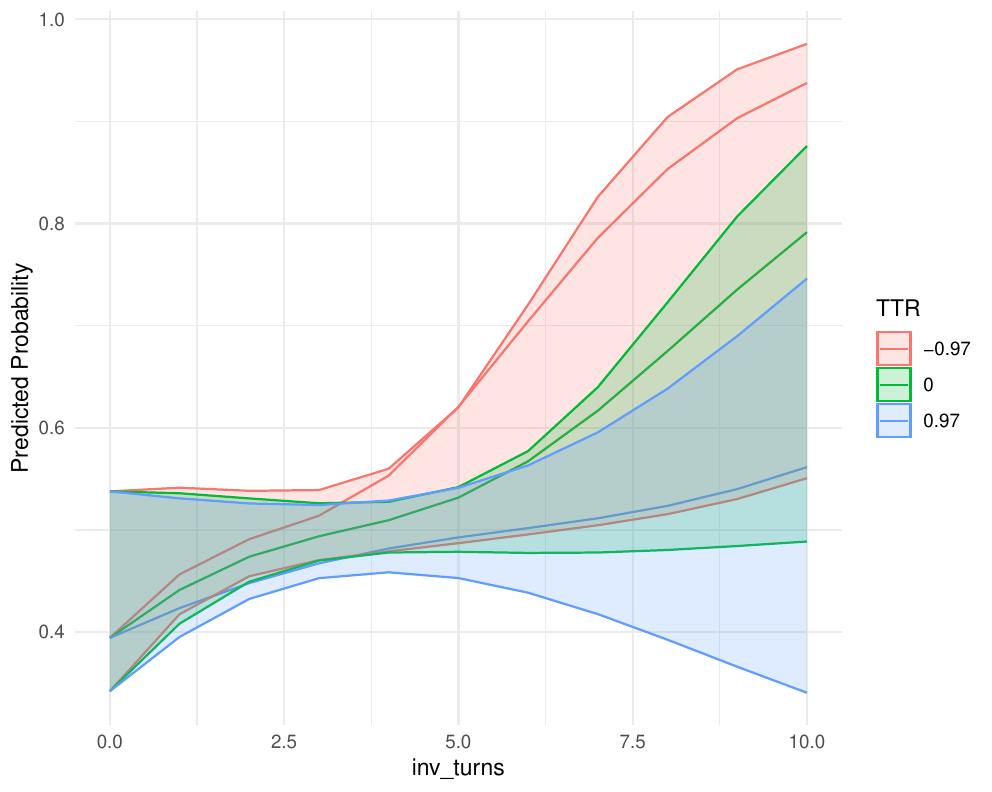}
  \label{fig:ttr}
}
\caption{The predicted values and confidence intervals of the interaction terms between linguistic markers and \texttt{inv}\_\texttt{turns}. The level of usage are denoted in color, where red indicates the lower usage (1 SD below the mean), green indicates the average usage, and blue indicates higher usage (1 SD above the mean). The x-axis indicates the number of turns from test administrators.}
\label{fig:effects}
\end{figure*}

\paragraph{The predicted effects of the interaction terms}
As shown in Figure~\ref{fig:pron}, we observed that $\mathcal{M}_{\text{pitt}}$ predicts a dramatic increase in the probability of having a dementia diagnosis from 0.1 to 0.8 as conversations went longer for participants who used lower level of pronoun during the test. For participants with average pronoun usage (at mean, shown in green), $\mathcal{M}_{\text{pitt}}$ maintained consistent predicted probabilities of having a dementia diagnosis throughout all conversation lengths. Conversely, participants with high pronoun usage showed an initial high probability of approximately 0.8 for have a dementia diagnosis in shorter conversations, which gradually decreased to 0.7 as conversation went longer. As we observed in Figure~\ref{fig:ttr}, participants with lower TTR (shown in red) had an increasing probability of having a dementia diagnosis as the number of turns from test administrators increased, rising dramatically from around 0.5 to nearly 0.95 over 10 turns. Notably, participants with higher TTR (shown in blue) showed a contrasting pattern - their probability of having a dementia diagnosis actually decreased slightly as conversations went longer, dropping from 0.5 to 0.35. Furthermore, we found that the predictive probabilities of pronoun usage and TTR varies systematically with  \texttt{inv}\_\texttt{turns}. Collectively, Figure~\ref{fig:effects} suggests an interesting diagnostic transition: at a lower level of test administrator involvement (\texttt{inv}\_\texttt{turns} $\leq$ 3, typical for healthy controls), pronoun usage provides greater diagnostic utility; at moderate involvement (\texttt{inv}\_\texttt{turns} $\approx$ 4, typical for matched dementia patients), both features offer complementary values; while at a higher involvement levels ((\texttt{inv}\_\texttt{turns} $\geq$ 6, typical for dementia patients before PSM), TTR becomes the dominant discriminative marker. This suggests that different linguistic features gain or lose diagnostic utility depending on the degree of administrator involvement.

\subsection{Cross validation: classification performance}

$\mathcal{M}_{\text{pitt}}$ achieved accuracy of 0.67, precision of 0.69, recall of 0.56, and F$_{1}$ score of 0.62 on the matched Pitt test split, respectively. Interestingly, $\mathcal{M}_{\text{pitt}}$ did not generalize well to the original WLS corpus, reaching accuracy of 0.59, precision of 0.13, recall of 0.38, and F$_{1}$ score of 0.19, respectively. $\mathcal{M}_{\text{pitt}}$ performed similarly on the matched WLS corpus, reaching accuracy of 0.50, precision of 0.50, recall of 0.38, and F$_{1}$ score of 0.43, respectively. $\mathcal{M}_{\text{wls}}$ also generalized poorly to the matched Pitt corpus, with accuracy of 0.55, precision of 0.54, recall of 0.47, and F$_{1}$ score of 0.50 on the matched Pitt test split.

\section{Discussion}

Our key findings are as follows. First, we show that many linguistic features previously studied in AD dementia progression appear to vary with level of test administrator involvement. Second, the observed variability between two corpora underscores the importance of considering administrator behavior as a potential confounding variable in linguistic analyses of clinical populations. These findings collectively suggest that some of the linguistic features commonly observed in dementia patients may be affected by the data collection processes rather than cognitive decline alone.

The observation of interactive test administrator dynamics in the Pitt corpus is consistent with prior work \citep{menn1989cross, CAPLAN1998184}, which report that the test administrator needs to induce \textit{enough} propositional utterances from participants in the constrained task such as the ``Cookie Theft'' picture description task. However, it is often noted that dementia patients are incapable of producing complex utterances due to the progression of the disease. As such, an interactive test administrator dynamic may lead to overestimation of a patients' linguistic ability in some cases.


Our results further suggest that these interaction patterns influence downstream dementia classification, which is consistent with a prior work \citep{farzana-parde-2022-interaction}. Our study further quantifies the influence of test administrator behavior, demonstrating how the varying levels of investigator involvement between groups may confound our interpretation of linguistic markers as diagnostic indicators. Our results highlight the need to interpret linguistic markers not as isolated indicators, but as features embedded within an interactive context that includes test administrators' role in shaping the discourse. Further research design might benefit from explicitly accounting for and potentially controlling test administrator involvement when developing screening criteria based on linguistic features.


Our findings suggest a nuanced relationship between linguistic markers, administrator interaction patterns, and their predictability for cognitive decline. The consistently high predicted probability of a higher probability of developing dementia for participants with elevated pronoun usage (shown by the stable high probabilities in the blue line in Figure~\ref{fig:pron}) supports existing literature on pronoun over-usage \citep{almor1999alzheimer, jarrold-etal-2014-aided, cummings2019describing} as a linguistic marker of cognitive decline. However, our results also indicate that this relationship may be masked or amplified by test administrators' interaction styles, as evidenced by varying predicted probability trajectories across different conversation lengths. Similarly, while the observed TTR patterns also align with previous findings \citep{hier1985language} that lower lexical diversity indicates cognitive decline, the dramatic increase in predicted probability for participants with lower TTR during longer conversations suggests that the established observations might be influenced by the test administrators' interaction patterns, suggests that these established linguistic markers may be partially attributable to differences in the test administration practices rather than the true construct measures of cognitive decline.

The disparities of classification performance of two models -- $\mathcal{M}_{\text{pitt}}$ for \textit{detecting} AD dementia, and $\mathcal{M}_{\text{wls}}$ for \textit{predicting} dementia -- confirms the often-observed challenges of developing robust and generalizable models for dementia detection and prediction. While $\mathcal{M}_{\text{pitt}}$ demonstrated moderate performance on its test split, it generalized poorly on the WLS corpus where precision and F$_{1}$ score dropped dramatically. $\mathcal{M}_{\text{pitt}}$'s slight improvement in performance on the matched vs. original WLS corpus suggests that the PSM may somewhat mitigate the confounding effect, but not fully resolve the cross-corpus and cross-task generalization issues. Similarly, $\mathcal{M}_{\text{wls}}$ showed limited generalization on the Pitt corpus. This consistent under-performance across corpora suggests the significant challenge of creating models that can \textit{reliably} detect or predict dementia. Our results also suggest the need of considering corpus- and population-specific characteristics in the model development. Factors such as demographic differences, test administrating styles, and the temporal aspect of dementia progression (i.e., detection vs. prediction) may contribute to the observed lack of cross-corpus and cross-task generalizability.

The variability between two corpora suggests that some linguistic markers previously attributed to dementia may be specific to certain data collection protocols rather than universal linguistic anomalies associated with the disease's progression. $\mathcal{M}_{\text{pitt}}$ demonstrates reasonable performance on its own test split, suggesting that within a single dataset, certain linguistic patterns may indeed be indicative of cognitive decline after controlling for the influence of test administrators. However, its substantially degraded performance on the WLS corpus points out a critical issue: linguistic markers that appear robust within one population may not translate effectively to another. This lack of cross-corpus generalizability persists when we validate $\mathcal{M}_{\text{wls}}$ on the Pitt corpus - the performance of $\mathcal{M}_{\text{wls}}$ actually worsens on the matched Pitt test split. These findings collectively suggest that the linguistic anomalies associated with AD progression may be highly context-dependent, influenced by factors such as data collection protocols, test administrator dynamics, and population-specific characteristics. This indicates the need for caution when interpreting linguistic markers of cognitive decline, developing specialized neural language models, and validating findings across diverse datasets and populations.

While the speech samples produced by population with high clinical risks are scarce, incorporating text corpora drawn from different sources (also known as confounding by provenance) presents both opportunities and challenges for detecting linguistic anomalies in AD dementia. Previous studies demonstrate that treating the provenance of a transcript (i.e., Pitt vs. WLS) as a secondary target for prediction \citep{guo2021crossing} and data augmentation \citep[\textit{inter alia}]{9413566, BERTINI2022101298, duan-etal-2023-cda} could lead to performance improvements. However, our results suggest the need for extra caution in such applications. These disparities suggest these approaches, if not carefully implemented, may introduce additional confounding variables rather than identifying true indicators of cognitive impairment. As such, the observed lack of cross-corpus and cross-task generalizability may explain why fine-tuned neural language models generalize less-than-ideal to other speech samples produced by populations at high clinical risk \citep{li-etal-2022-gpt, farzana-parde-2023-towards}.

While the automated analysis of spoken language produced by population with high clinical risk remains a valuable component of early-screening cognitive assessment, the observed influence of test administrator dynamics on AD-related linguistic anomalies calls for a re-evaluation of current methods. Researchers and clinicians should exercise caution when interpreting the linguistic features of the ``Cookie Theft'' picture description task, as they may be partially artifacts of the data collection itself. Our results call for a standardized test administration to minimize the variability in administrator engagement, and the need for population- and language-specific norms for assessments. 

\section{Conclusion}

Our study explored the relationship between test administrator involvement and linguistic features in dementia assessments using the ``Cookie Theft'' picture description task. The patterns we observed raise questions about how established linguistic features might be shaped by the dynamics of test administration alongside cognitive status. Our study brings the potential benefits of considering administrator behavior in future development of clinical speech analytics frameworks.


\newpage
\section*{Limitations}
The work presented here has several limitations. While our analysis identifies significant correlations between the test administrator interactions and linguistic features, we should note that our study design does not establish a direct causal link. Future experimental studies with standardized administrator protocols would be necessary to establish such a link. Second, the size of the datasets used in this study is considerably small, which is a common concern in this line of research \citep{petti2020systematic}. Moreover, all datasets used in this study are in American English, and many participants are representative of White, non-Hispanic American residents, which certainly limits the generalizability to other languages and ethnic groups. In this study, we only focus on analyzing POS tags for both datasets, which is a limited feature set for detecting cognitive impairment. Future studies should explore comprehensive linguistic and acoustic features (i.e., \citet{fraser2015linguistic}) to establish a more definitive measurement of the effects of test administrator engagement. We acknowledge that there are linguistic differences between the two corpora studied in this work \citep{johnstone2015pittsburgh}, which may affect the comparability of results across datasets. We should also note that while category fluency task has demonstrates the clinical utility for dementia screening; it is, however, not a complete clinical diagnosis, which may not capture the full spectrum of cognitive decline and could potentially lead to misclassification of some participants.

\section*{Acknowledgement}
Research reported in this publication was supported by the National Library of Medicine of the National Institutes of Health under award number R01 LM014056-02S1, and the National Institute on Aging of the National institutes of Health under award number AG069792.

\bibliography{anthology,custom}

\begin{thebibliography}{60}
\providecommand{\natexlab}[1]{#1}

\bibitem[{Akaike(1998)}]{akaike1998information}
Hirotogu Akaike. 1998.
\newblock Information theory and an extension of the maximum likelihood
  principle.
\newblock In \emph{Selected papers of hirotugu akaike}, pages 199--213.
  Springer.

\bibitem[{Almor et~al.(1999)Almor, Kempler, MacDonald, Andersen, and
  Tyler}]{almor1999alzheimer}
Amit Almor, Daniel Kempler, Maryellen~C MacDonald, Elaine~S Andersen, and
  Lorraine~K Tyler. 1999.
\newblock Why do alzheimer patients have difficulty with pronouns? working
  memory, semantics, and reference in comprehension and production in
  alzheimer's disease.
\newblock \emph{Brain and language}, 67(3):202--227.

\bibitem[{Ash et~al.(2006)Ash, Moore, Antani, McCawley, Work, and
  Grossman}]{doi:10.1212/01.wnl.0000210435.72614.38}
S.~Ash, P.~Moore, S.~Antani, G.~McCawley, M.~Work, and M.~Grossman. 2006.
\newblock \href {https://doi.org/10.1212/01.wnl.0000210435.72614.38} {Trying to
  tell a tale}.
\newblock \emph{Neurology}, 66(9):1405--1413.

\bibitem[{Austin(2011)}]{doi:10.1080/00273171.2011.568786}
Peter~C. Austin. 2011.
\newblock \href {https://doi.org/10.1080/00273171.2011.568786} {An introduction
  to propensity score methods for reducing the effects of confounding in
  observational studies}.
\newblock \emph{Multivariate Behavioral Research}, 46(3):399--424.

\bibitem[{Banovic et~al.(2018)Banovic, Zunic, and
  Sinanovic}]{banovic2018communication}
Silva Banovic, Lejla~Junuzovic Zunic, and Osman Sinanovic. 2018.
\newblock Communication difficulties as a result of dementia.
\newblock \emph{Materia socio-medica}, 30(3):221.

\bibitem[{Becker et~al.(1994)Becker, Boiler, Lopez, Saxton, and
  McGonigle}]{10.1001/archneur.1994.00540180063015}
James~T. Becker, François Boiler, Oscar~L. Lopez, Judith Saxton, and Karen~L.
  McGonigle. 1994.
\newblock \href {https://doi.org/10.1001/archneur.1994.00540180063015} {{The
  Natural History of Alzheimer's Disease: Description of Study Cohort and
  Accuracy of Diagnosis}}.
\newblock \emph{Archives of Neurology}, 51(6):585--594.

\bibitem[{Bertini et~al.(2022)Bertini, Allevi, Lutero, Calzà, and
  Montesi}]{BERTINI2022101298}
Flavio Bertini, Davide Allevi, Gianluca Lutero, Laura Calzà, and Danilo
  Montesi. 2022.
\newblock \href {https://doi.org/10.1016/j.csl.2021.101298} {An automatic
  alzheimer’s disease classifier based on spontaneous spoken english}.
\newblock \emph{Computer Speech \& Language}, 72:101298.

\bibitem[{Berube et~al.(2019)Berube, Nonnemacher, Demsky, Glenn, Saxena,
  Wright, Tippett, and Hillis}]{berube2019stealing}
Shauna Berube, Jodi Nonnemacher, Cornelia Demsky, Shenly Glenn, Sadhvi Saxena,
  Amy Wright, Donna~C Tippett, and Argye~E Hillis. 2019.
\newblock Stealing cookies in the twenty-first century: Measures of spoken
  narrative in healthy versus speakers with aphasia.
\newblock \emph{American journal of speech-language pathology},
  28(1S):321--329.

\bibitem[{Blanken et~al.(1987)Blanken, Dittmann, Haas, and
  Wallesch}]{blanken1987spontaneous}
Gerhard Blanken, J{\"u}rgen Dittmann, J-Christian Haas, and Claus-W Wallesch.
  1987.
\newblock Spontaneous speech in senile dementia and aphasia: Implications for a
  neurolinguistic model of language production.
\newblock \emph{Cognition}, 27(3):247--274.

\bibitem[{Brysbaert and New(2009)}]{brysbaert2009moving}
Marc Brysbaert and Boris New. 2009.
\newblock \href {https://doi.org/10.3758/BRM.41.4.977} {Moving beyond
  ku{\v{c}}era and francis: A critical evaluation of current word frequency
  norms and the introduction of a new and improved word frequency measure for
  american english}.
\newblock \emph{Behavior research methods}, 41(4):977--990.

\bibitem[{Bucks et~al.(2000)Bucks, Singh, Cuerden, and
  Wilcock}]{bucks2000analysis}
Romola~S Bucks, Sameer Singh, Joanne~M Cuerden, and Gordon~K Wilcock. 2000.
\newblock Analysis of spontaneous, conversational speech in dementia of
  alzheimer type: Evaluation of an objective technique for analysing lexical
  performance.
\newblock \emph{Aphasiology}, 14(1):71--91.

\bibitem[{Caama{\~n}o-Isorna et~al.(2006)Caama{\~n}o-Isorna, Corral,
  Montes-Mart{\'\i}nez, and Takkouche}]{caamano2006education}
Francisco Caama{\~n}o-Isorna, Montserrat Corral, Agust{\'\i}n
  Montes-Mart{\'\i}nez, and Bahi Takkouche. 2006.
\newblock \href {https://doi.org/10.1159/000093378} {Education and dementia: a
  meta-analytic study}.

\bibitem[{Canning et~al.(2004)Canning, Leach, Stuss, Ngo, and
  Black}]{canning2004diagnostic}
SJ~Duff Canning, L~Leach, D~Stuss, L~Ngo, and SE14981170 Black. 2004.
\newblock Diagnostic utility of abbreviated fluency measures in alzheimer
  disease and vascular dementia.
\newblock \emph{Neurology}, 62(4):556--562.

\bibitem[{Caplan and Hanna(1998)}]{CAPLAN1998184}
David Caplan and Joy~E. Hanna. 1998.
\newblock \href {https://doi.org/10.1006/brln.1998.1930} {Sentence production
  by aphasic patients in a constrained task}.
\newblock \emph{Brain and Language}, 63(2):184--218.

\bibitem[{Cerhan et~al.(2002)Cerhan, Ivnik, Smith, Tangalos, Petersen, and
  Boeve}]{cerhan2002diagnostic}
Jane~H Cerhan, Robert~J Ivnik, Glenn~E Smith, Eric~C Tangalos, Ronald~C
  Petersen, and Bradley~F Boeve. 2002.
\newblock Diagnostic utility of letter fluency, category fluency, and fluency
  difference scores in alzheimer's disease.
\newblock \emph{The Clinical Neuropsychologist}, 16(1):35--42.

\bibitem[{Chandler et~al.(2019)Chandler, Cumpston, Li, Page, and
  Welch}]{chandler2019cochrane}
Jacqueline Chandler, Miranda Cumpston, Tianjing Li, Matthew~J Page, and VJHW
  Welch. 2019.
\newblock Cochrane handbook for systematic reviews of interventions.
\newblock \emph{Hoboken: Wiley}.

\bibitem[{Crockford and Lesser(1994)}]{crockford1994assessing}
Catherine Crockford and Ruth Lesser. 1994.
\newblock Assessing functional communication in aphasia: Clinical utility and
  time demands of three methods.
\newblock \emph{International Journal of Language \& Communication Disorders},
  29(2):165--182.

\bibitem[{Cummings(2019)}]{cummings2019describing}
Louise Cummings. 2019.
\newblock Describing the cookie theft picture: Sources of breakdown in
  alzheimer’s dementia.
\newblock \emph{Pragmatics and Society}, 10(2):153--176.

\bibitem[{Custodio et~al.(2020)Custodio, Duque, Montesinos, Alva-Diaz, Mellado,
  and Slachevsky}]{10.3389/fnagi.2020.00270}
Nilton Custodio, Lissette Duque, Rosa Montesinos, Carlos Alva-Diaz, Martin
  Mellado, and Andrea Slachevsky. 2020.
\newblock \href {https://doi.org/10.3389/fnagi.2020.00270} {Systematic review
  of the diagnostic validity of brief cognitive screenings for early dementia
  detection in spanish-speaking adults in latin america}.
\newblock \emph{Frontiers in Aging Neuroscience}, 12.

\bibitem[{Ding et~al.(2024)Ding, Chetty, Noori~Hoshyar, Bhattacharya, and
  Klein}]{ding2024speech}
Kewen Ding, Madhu Chetty, Azadeh Noori~Hoshyar, Tanusri Bhattacharya, and Britt
  Klein. 2024.
\newblock \href {https://doi.org/10.1007/s10462-024-10961-6} {Speech based
  detection of alzheimer’s disease: a survey of ai techniques, datasets and
  challenges}.
\newblock \emph{Artificial Intelligence Review}, 57(12):1--43.

\bibitem[{Duan et~al.(2023)Duan, Wei, Liu, Li, Liu, and
  Wang}]{duan-etal-2023-cda}
Junwen Duan, Fangyuan Wei, Jin Liu, Hongdong Li, Tianming Liu, and Jianxin
  Wang. 2023.
\newblock \href {https://doi.org/10.18653/v1/2023.findings-acl.114} {{CDA}: A
  contrastive data augmentation method for {A}lzheimer{'}s disease detection}.
\newblock In \emph{Findings of the Association for Computational Linguistics:
  ACL 2023}, pages 1819--1826, Toronto, Canada. Association for Computational
  Linguistics.

\bibitem[{Eggenberger et~al.(2012)Eggenberger, Heimerl, and
  Bennett}]{Eggenberger2012CommunicationST}
Eva Eggenberger, Katharina Heimerl, and Michael~I. Bennett. 2012.
\newblock \href {https://api.semanticscholar.org/CorpusID:28386310}
  {Communication skills training in dementia care: a systematic review of
  effectiveness, training content, and didactic methods in different care
  settings}.
\newblock \emph{International Psychogeriatrics}, 25:345 -- 358.

\bibitem[{Farzana and Parde(2022)}]{farzana-parde-2022-interaction}
Shahla Farzana and Natalie Parde. 2022.
\newblock \href {https://doi.org/10.18653/v1/2022.sigdial-1.18} {Are
  interaction patterns helpful for task-agnostic dementia detection? an
  empirical exploration}.
\newblock In \emph{Proceedings of the 23rd Annual Meeting of the Special
  Interest Group on Discourse and Dialogue}, pages 172--182, Edinburgh, UK.
  Association for Computational Linguistics.

\bibitem[{Farzana and Parde(2023)}]{farzana-parde-2023-towards}
Shahla Farzana and Natalie Parde. 2023.
\newblock \href {https://doi.org/10.18653/v1/2023.acl-long.668} {Towards
  domain-agnostic and domain-adaptive dementia detection from spoken language}.
\newblock In \emph{Proceedings of the 61st Annual Meeting of the Association
  for Computational Linguistics (Volume 1: Long Papers)}, pages 11965--11978,
  Toronto, Canada. Association for Computational Linguistics.

\bibitem[{Fraser et~al.(2015)Fraser, Meltzer, and
  Rudzicz}]{fraser2015linguistic}
Kathleen~C Fraser, Jed~A Meltzer, and Frank Rudzicz. 2015.
\newblock \href {https://doi.org/10.3233/JAD-150520} {Linguistic features
  identify alzheimer’s disease in narrative speech}.
\newblock \emph{Journal of Alzheimer’s disease}, 49(2):407--422.

\bibitem[{Gewirth et~al.(1984)Gewirth, Shindler, and Hier}]{gewirth1984altered}
Letitia~R Gewirth, Andrea~G Shindler, and Daniel~B Hier. 1984.
\newblock Altered patterns of word associations in dementia and aphasia.
\newblock \emph{Brain and Language}, 21(2):307--317.

\bibitem[{Goodglass and Kaplan(1983)}]{goodglass1983boston}
Harold Goodglass and Edith Kaplan. 1983.
\newblock \emph{Boston diagnostic aphasia examination booklet}.
\newblock Lea \& Febiger.

\bibitem[{Gumus et~al.(2024)Gumus, Koo, Studzinski, Bhan, Robin, and
  Black}]{Gumus2024LinguisticCI}
Melisa Gumus, Morgan Koo, Christa~M. Studzinski, Aparna Bhan, Jessica Robin,
  and Sandra~E. Black. 2024.
\newblock \href {https://api.semanticscholar.org/CorpusID:268710857}
  {Linguistic changes in neurodegenerative diseases relate to clinical
  symptoms}.
\newblock \emph{Frontiers in Neurology}, 15.

\bibitem[{Guo et~al.(2021)Guo, Li, Roan, Pakhomov, and Cohen}]{guo2021crossing}
Yue Guo, Changye Li, Carol Roan, Serguei Pakhomov, and Trevor Cohen. 2021.
\newblock \href {https://doi.org/10.3389/fcomp.2021.642517} {Crossing the
  “cookie theft” corpus chasm: applying what bert learns from outside data
  to the adress challenge dementia detection task}.
\newblock \emph{Frontiers in Computer Science}, 3:642517.

\bibitem[{Herd et~al.(2014)Herd, Carr, and Roan}]{10.1093/ije/dys194}
Pamela Herd, Deborah Carr, and Carol Roan. 2014.
\newblock \href {https://doi.org/10.1093/ije/dys194} {{Cohort Profile:
  Wisconsin longitudinal study (WLS)}}.
\newblock \emph{International Journal of Epidemiology}, 43(1):34--41.

\bibitem[{Hier et~al.(1985)Hier, Hagenlocker, and Shindler}]{hier1985language}
Daniel~B Hier, Karen Hagenlocker, and Andrea~Gellin Shindler. 1985.
\newblock Language disintegration in dementia: Effects of etiology and
  severity.
\newblock \emph{Brain and language}, 25(1):117--133.

\bibitem[{Jarrold et~al.(2014)Jarrold, Peintner, Wilkins, Vergryi, Richey,
  Gorno-Tempini, and Ogar}]{jarrold-etal-2014-aided}
William Jarrold, Bart Peintner, David Wilkins, Dimitra Vergryi, Colleen Richey,
  Maria~Luisa Gorno-Tempini, and Jennifer Ogar. 2014.
\newblock \href {https://doi.org/10.3115/v1/W14-3204} {Aided diagnosis of
  dementia type through computer-based analysis of spontaneous speech}.
\newblock In \emph{Proceedings of the Workshop on Computational Linguistics and
  Clinical Psychology: From Linguistic Signal to Clinical Reality}, pages
  27--37, Baltimore, Maryland, USA. Association for Computational Linguistics.

\bibitem[{Johnstone et~al.(2015)Johnstone, Baumgardt, Eberhardt, and
  Kiesling}]{johnstone2015pittsburgh}
Barbara Johnstone, Daniel Baumgardt, Maeve Eberhardt, and Scott Kiesling. 2015.
\newblock \emph{Pittsburgh speech and Pittsburghese}, volume~11.
\newblock Walter de Gruyter GmbH \& Co KG.

\bibitem[{Labov(1973)}]{labov1973sociolinguistic}
William Labov. 1973.
\newblock \emph{Sociolinguistic patterns}.
\newblock 4. University of Pennsylvania press.

\bibitem[{Li et~al.(2022)Li, Knopman, Xu, Cohen, and
  Pakhomov}]{li-etal-2022-gpt}
Changye Li, David Knopman, Weizhe Xu, Trevor Cohen, and Serguei Pakhomov. 2022.
\newblock \href {https://doi.org/10.18653/v1/2022.acl-long.131} {{GPT}-{D}:
  Inducing dementia-related linguistic anomalies by deliberate degradation of
  artificial neural language models}.
\newblock In \emph{Proceedings of the 60th Annual Meeting of the Association
  for Computational Linguistics (Volume 1: Long Papers)}, pages 1866--1877,
  Dublin, Ireland. Association for Computational Linguistics.

\bibitem[{Li et~al.(2023)Li, Xu, Cohen, Michalowski, and
  Pakhomov}]{li2023trestle}
Changye Li, Weizhe Xu, Trevor Cohen, Martin Michalowski, and Serguei Pakhomov.
  2023.
\newblock Trestle: Toolkit for reproducible execution of speech, text and
  language experiments.
\newblock \emph{AMIA Summits on Translational Science Proceedings}, 2023:360.

\bibitem[{Liu et~al.(2019)Liu, Ott, Goyal, Du, Joshi, Chen, Levy, Lewis,
  Zettlemoyer, and Stoyanov}]{Liu2019RoBERTaAR}
Yinhan Liu, Myle Ott, Naman Goyal, Jingfei Du, Mandar Joshi, Danqi Chen, Omer
  Levy, Mike Lewis, Luke Zettlemoyer, and Veselin Stoyanov. 2019.
\newblock \href {https://api.semanticscholar.org/CorpusID:198953378} {Roberta:
  A robustly optimized bert pretraining approach}.
\newblock \emph{ArXiv}, abs/1907.11692.

\bibitem[{Liu et~al.(2021)Liu, Guo, Ling, and Li}]{9413566}
Zhaoci Liu, Zhiqiang Guo, Zhenhua Ling, and Yunxia Li. 2021.
\newblock \href {https://doi.org/10.1109/ICASSP39728.2021.9413566} {Detecting
  alzheimer’s disease from speech using neural networks with bottleneck
  features and data augmentation}.
\newblock In \emph{ICASSP 2021 - 2021 IEEE International Conference on
  Acoustics, Speech and Signal Processing (ICASSP)}, pages 7323--7327.

\bibitem[{Luz et~al.(2020)Luz, Haider, de~la Fuente, Fromm, and
  MacWhinney}]{luz20_interspeech}
Saturnino Luz, Fasih Haider, Sofia de~la Fuente, Davida Fromm, and Brian
  MacWhinney. 2020.
\newblock \href {https://doi.org/10.21437/Interspeech.2020-2571}
  {{Alzheimer’s Dementia Recognition Through Spontaneous Speech: The ADReSS
  Challenge}}.
\newblock In \emph{Proc. Interspeech 2020}, pages 2172--2176.

\bibitem[{Menn and Obler(1989)}]{menn1989cross}
Lise Menn and Loraine~K Obler. 1989.
\newblock Cross-language data and theories of agrammatism.
\newblock In \emph{Agrammatic aphasia}, pages 1369--1389. John Benjamins.

\bibitem[{Monsch et~al.(1992)Monsch, Bondi, Butters, Salmon, Katzman, and
  Thal}]{monsch1992comparisons}
Andreas~U Monsch, Mark~W Bondi, Nelson Butters, David~P Salmon, Robert Katzman,
  and Leon~J Thal. 1992.
\newblock Comparisons of verbal fluency tasks in the detection of dementia of
  the alzheimer type.
\newblock \emph{Archives of neurology}, 49(12):1253--1258.

\bibitem[{Ngandu et~al.(2007)Ngandu, von Strauss, Helkala, Winblad, Nissinen,
  Tuomilehto, Soininen, and Kivipelto}]{ngandu2007education}
Tiia Ngandu, Eva von Strauss, E-L Helkala, B~Winblad, A~Nissinen, J~Tuomilehto,
  H~Soininen, and M~Kivipelto. 2007.
\newblock \href {https://doi.org/10.1212/01.wnl.0000277456.29440.16} {Education
  and dementia: what lies behind the association?}
\newblock \emph{Neurology}, 69(14):1442--1450.

\bibitem[{Nguyen et~al.(2016)Nguyen, Tchetgen, Kawachi, Gilman, Walter, Liu,
  Manly, and Glymour}]{nguyen2016instrumental}
Thu~T Nguyen, Eric J~Tchetgen Tchetgen, Ichiro Kawachi, Stephen~E Gilman,
  Stefan Walter, Sze~Y Liu, Jennifer~J Manly, and M~Maria Glymour. 2016.
\newblock \href {https://doi.org/10.1016/j.annepidem.2015.10.006} {Instrumental
  variable approaches to identifying the causal effect of educational
  attainment on dementia risk}.
\newblock \emph{Annals of epidemiology}, 26(1):71--76.

\bibitem[{Nicholas et~al.(1985)Nicholas, Obler, Albert, and nancy~helm
  estabrooks}]{Nicholas1985EmptySI}
Marjorie Nicholas, Loraine~K. Obler, Martin~L. Albert, and nancy~helm
  estabrooks. 1985.
\newblock \href {https://api.semanticscholar.org/CorpusID:43387831} {Empty
  speech in alzheimer's disease and fluent aphasia.}
\newblock \emph{Journal of speech and hearing research}, 28 3:405--10.

\bibitem[{Pakhomov et~al.(2011)Pakhomov, Chacon, Wicklund, and
  Gundel}]{pakhomov2011computerized}
Serguei Pakhomov, Dustin Chacon, Mark Wicklund, and Jeanette Gundel. 2011.
\newblock \href {https://doi.org/10.3758/s13428-010-0037-9} {Computerized
  assessment of syntactic complexity in alzheimer’s disease: a case study of
  iris murdoch’s writing}.
\newblock \emph{Behavior research methods}, 43:136--144.

\bibitem[{Petti et~al.(2020)Petti, Baker, and Korhonen}]{petti2020systematic}
Ulla Petti, Simon Baker, and Anna Korhonen. 2020.
\newblock \href {https://doi.org/10.1093/jamia/ocaa174} {A systematic
  literature review of automatic alzheimer’s disease detection from speech
  and language}.
\newblock \emph{Journal of the American Medical Informatics Association},
  27(11):1784--1797.

\bibitem[{Petti et~al.(2023)Petti, Baker, Korhonen, and
  Robin}]{10.1159/000533423}
Ulla Petti, Simon Baker, Anna Korhonen, and Jessica Robin. 2023.
\newblock \href {https://doi.org/10.1159/000533423} {{How Much Speech Data Is
  Needed for Tracking Language Change in Alzheimer’s Disease? A Comparison of
  Random Length, 5-Min, and 1-Min Spontaneous Speech Samples}}.
\newblock \emph{Digital Biomarkers}, 7(1):157--166.

\bibitem[{Reddy(2022)}]{reddy2022explainability}
Sandeep Reddy. 2022.
\newblock Explainability and artificial intelligence in medicine.
\newblock \emph{The Lancet Digital Health}, 4(4):e214--e215.

\bibitem[{Rousseaux et~al.(2010{\natexlab{a}})Rousseaux, S{\`e}ve, Vallet,
  Pasquier, and Mackowiak-Cordoliani}]{rousseaux2010analysis}
Marc Rousseaux, Amandine S{\`e}ve, Marion Vallet, Florence Pasquier, and
  Marie~Anne Mackowiak-Cordoliani. 2010{\natexlab{a}}.
\newblock An analysis of communication in conversation in patients with
  dementia.
\newblock \emph{Neuropsychologia}, 48(13):3884--3890.

\bibitem[{Rousseaux et~al.(2010{\natexlab{b}})Rousseaux, Sève, Vallet,
  Pasquier, and Mackowiak-Cordoliani}]{ROUSSEAUX20103884}
Marc Rousseaux, Amandine Sève, Marion Vallet, Florence Pasquier, and
  Marie~Anne Mackowiak-Cordoliani. 2010{\natexlab{b}}.
\newblock \href {https://doi.org/10.1016/j.neuropsychologia.2010.09.026} {An
  analysis of communication in conversation in patients with dementia}.
\newblock \emph{Neuropsychologia}, 48(13):3884--3890.

\bibitem[{Ruitenberg et~al.(2001)Ruitenberg, Ott, {van Swieten}, Hofman, and
  Breteler}]{RUITENBERG2001575}
Annemieke Ruitenberg, Alewijn Ott, John~C. {van Swieten}, Albert Hofman, and
  Monique~M.B. Breteler. 2001.
\newblock \href {https://doi.org/10.1016/S0197-4580(01)00231-7} {Incidence of
  dementia: does gender make a difference?}
\newblock \emph{Neurobiology of Aging}, 22(4):575--580.

\bibitem[{Sabat(1994)}]{sabat1994language}
Steven~R Sabat. 1994.
\newblock Language function in alzheimer's disease: a critical review of
  selected literature.
\newblock \emph{Language \& Communication}, 14(4):331--351.

\bibitem[{Seyed Ahmad~Sajjadi and
  Nestor(2012)}]{doi:10.1080/02687038.2012.654933}
Michal~Tomek Seyed Ahmad~Sajjadi, Karalyn~Patterson and Peter~J. Nestor. 2012.
\newblock \href {https://doi.org/10.1080/02687038.2012.654933} {Abnormalities
  of connected speech in semantic dementia vs alzheimer's disease}.
\newblock \emph{Aphasiology}, 26(6):847--866.

\bibitem[{Shi et~al.(2023)Shi, Cheung, and Shahamiri}]{SHI2023115538}
Mengke Shi, Gary Cheung, and Seyed~Reza Shahamiri. 2023.
\newblock \href {https://doi.org/10.1016/j.psychres.2023.115538} {Speech and
  language processing with deep learning for dementia diagnosis: A systematic
  review}.
\newblock \emph{Psychiatry Research}, 329:115538.

\bibitem[{Snowdon et~al.(1996)Snowdon, Kemper, Mortimer, Greiner, Wekstein, and
  Markesbery}]{snowdon1996linguistic}
David~A Snowdon, Susan~J Kemper, James~A Mortimer, Lydia~H Greiner, David~R
  Wekstein, and William~R Markesbery. 1996.
\newblock \href {https://doi.org/10.1001/jama.1996.03530310034029} {Linguistic
  ability in early life and cognitive function and alzheimer's disease in late
  life: Findings from the nun study}.
\newblock \emph{Jama}, 275(7):528--532.

\bibitem[{Stokes et~al.(2015)Stokes, Combes, and
  Stokes}]{doi.org/10.1111/psyg.12095}
Laura Stokes, Helen Combes, and Graham Stokes. 2015.
\newblock \href {https://doi.org/10.1111/psyg.12095} {The dementia diagnosis: a
  literature review of information, understanding, and attributions}.
\newblock \emph{Psychogeriatrics}, 15(3):218--225.

\bibitem[{van~der Flier and Scheltens(2005)}]{van2005epidemiology}
Wiesje~M van~der Flier and Philip Scheltens. 2005.
\newblock \href {https://doi.org/10.1136/jnnp.2005.082867} {Epidemiology and
  risk factors of dementia}.
\newblock \emph{Journal of Neurology, Neurosurgery \& Psychiatry}, 76(suppl
  5):v2--v7.

\bibitem[{Vaswani et~al.(2017)Vaswani, Shazeer, Parmar, Uszkoreit, Jones,
  Gomez, Kaiser, and Polosukhin}]{NIPS2017_3f5ee243}
Ashish Vaswani, Noam Shazeer, Niki Parmar, Jakob Uszkoreit, Llion Jones,
  Aidan~N Gomez, \L~ukasz Kaiser, and Illia Polosukhin. 2017.
\newblock \href
  {https://proceedings.neurips.cc/paper_files/paper/2017/file/3f5ee243547dee91fbd053c1c4a845aa-Paper.pdf}
  {Attention is all you need}.
\newblock In \emph{Advances in Neural Information Processing Systems},
  volume~30. Curran Associates, Inc.

\bibitem[{Weiner et~al.(2016)Weiner, Herff, and Schultz}]{weiner16_interspeech}
Jochen Weiner, Christian Herff, and Tanja Schultz. 2016.
\newblock \href {https://doi.org/10.21437/Interspeech.2016-100} {Speech-based
  detection of alzheimer’s disease in conversational german}.
\newblock In \emph{Interspeech 2016}, pages 1938--1942.

\bibitem[{Zhang et~al.(2019)Zhang, Kim, Lonjon, Zhu et~al.}]{zhang2019balance}
Zhongheng Zhang, Hwa~Jung Kim, Guillaume Lonjon, Yibing Zhu, et~al. 2019.
\newblock \href {https://doi.org/10.21037/atm.2018.12.10} {Balance diagnostics
  after propensity score matching}.
\newblock \emph{Annals of translational medicine}, 7(1).

\end{thebibliography}

\section*{Appendix}
\label{sec:appendix}

\begin{table}[ht]
\centering
\small
\begin{tabular}{@{}|c|p{2cm}|@{}}
\toprule
\textbf{POS tags} & \textbf{Name} \\ \midrule
ADJ & Adjective \\ \midrule
ADP & Adposition\\ \midrule
ADV & Adverb \\ \midrule
AUX & Auxiliary \\ \midrule
CCONJ &\RaggedRight Coordinating conjunction \\ \midrule
DET & Determiner \\ \midrule
INTJ & Interjection \\ \midrule
NOUN & Noun\\ \midrule
PART & Particle \\ \midrule
PRON & Pronoun \\ \midrule
PROPN & Proper noun\\ \midrule
SCONJ & subordinating conjection \\ \midrule
VERB & Verb \\ \bottomrule
\end{tabular}
\caption{The Universal POS tags}
\label{tab:pos}
\end{table}

\begin{table*}[ht]
\centering
\small
\resizebox{0.7\linewidth}{!}{%
\begin{tabular}{@{}|ccclc|cccc|@{}}
\toprule
\multicolumn{1}{|c|}{\multirow{2}{*}{\textbf{Features}}} &
  \multicolumn{4}{c|}{\textbf{\textit{Before matching}}} &
  \multicolumn{4}{c|}{\textbf{\textit{After matching}}} \\ \cmidrule(l){2-9} 
\multicolumn{1}{|c|}{} &
  \multicolumn{1}{c|}{Level} &
  \multicolumn{1}{c|}{Control} &
  \multicolumn{1}{c|}{Dementia} &
 SMD  &
  \multicolumn{1}{c|}{\multirow{2}{*}{Level}} &
  \multicolumn{1}{c|}{Control} &
  \multicolumn{1}{c|}{Dementia} & SMD
   \\ \cmidrule(r){1-5} \cmidrule(l){7-9} 
\multicolumn{2}{|c|}{Number of transcripts ($n$)} &
  \multicolumn{1}{c|}{182} &
  \multicolumn{1}{c|}{214} &
   &
  \multicolumn{1}{c|}{} &
  \multicolumn{1}{c|}{167} &
  \multicolumn{1}{c|}{167} &
   \\ \midrule
\multicolumn{2}{|c|}{Education (mean (SD))} &
  \multicolumn{1}{c|}{13.92 (2.42)} &
  \multicolumn{1}{c|}{12.28 (2.81)} & 0.629
   &
  \multicolumn{1}{c|}{} &
  \multicolumn{1}{c|}{13.66 (2.24)} &
  \multicolumn{1}{c|}{12.53 (2.93)} & 0.434
   \\ \cmidrule(r){1-5} \cmidrule(l){7-9} 
\multicolumn{2}{|c|}{Age (mean (SD))} &
  \multicolumn{1}{c|}{64.08 (7.91)} &
  \multicolumn{1}{c|}{71.51 (8.63)} & 0.897
   & 
  \multicolumn{1}{c|}{} &
  \multicolumn{1}{c|}{64.27 (7.85)} &
  \multicolumn{1}{c|}{71.46 (8.63)} & 0.871
   \\ \cmidrule(l){1-9} 
\multicolumn{1}{|c|}{\multirow{2}{*}{Gender (\%)}} &
  \multicolumn{1}{c|}{Female} &
  \multicolumn{1}{c|}{114 (62.6)} &
  \multicolumn{1}{c|}{147 (68.7)} & 0.128
   &
  \multicolumn{1}{c|}{Female} &
  \multicolumn{1}{c|}{104 (62.3)} &
  \multicolumn{1}{c|}{116 (69.5)} & 0.152
   \\ \cmidrule(l){2-9} 
\multicolumn{1}{|c|}{} &
  \multicolumn{1}{c|}{Male} &
  \multicolumn{1}{c|}{68 (37.4)} &
  \multicolumn{1}{c|}{67 (31.3)} &
   &
  \multicolumn{1}{c|}{Male} &
  \multicolumn{1}{c|}{63 (37.7)} &
  \multicolumn{1}{c|}{51 (30.5)} &
   \\ \midrule
\multicolumn{2}{|c|}{PRON (mean (SD))} &
  \multicolumn{1}{c|}{15.03 (9.85)} &
  \multicolumn{1}{c|}{17.18 (12.36)} & 0.193
   &
  \multicolumn{1}{c|}{\multirow{21}{*}{}} &
  \multicolumn{1}{c|}{14.72 (9.48)} &
  \multicolumn{1}{c|}{15.59 (10.80)} & 0.086
   \\ \cmidrule(r){1-5} \cmidrule(l){7-9} 
\multicolumn{2}{|c|}{PROPN (mean (SD))} &
  \multicolumn{1}{c|}{0.12 (0.51)} &
  \multicolumn{1}{c|}{0.25 (0.65)} & 0.227
   &
  \multicolumn{1}{c|}{} &
  \multicolumn{1}{c|}{0.13 (0.53)} &
  \multicolumn{1}{c|}{0.14 (0.46)} & 0.024
   \\ \cmidrule(r){1-5} \cmidrule(l){7-9} 
\multicolumn{2}{|c|}{NOUN (mean (SD))} &
  \multicolumn{1}{c|}{24.93 (13.94)} &
  \multicolumn{1}{c|}{19.41 (11.06)} & 0.439 
   &
  \multicolumn{1}{c|}{} &
  \multicolumn{1}{c|}{24.57 (13.96)} &
  \multicolumn{1}{c|}{19.37 (10.88)} & 0.416
   \\ \cmidrule(r){1-5} \cmidrule(l){7-9} 
\multicolumn{2}{|c|}{ADJ (mean (SD))} &
  \multicolumn{1}{c|}{4.06 (3.52)} &
  \multicolumn{1}{c|}{3.21 (3.47)} & 0.243
   &
  \multicolumn{1}{c|}{} &
  \multicolumn{1}{c|}{3.88 (3.38)} &
  \multicolumn{1}{c|}{3.16 (3.48)} & 0.211
   \\ \cmidrule(r){1-5} \cmidrule(l){7-9} 
\multicolumn{2}{|c|}{ADV (mean (SD))} &
  \multicolumn{1}{c|}{3.91 (3.75)} &
  \multicolumn{1}{c|}{5.43 (5.03)} & 0.342
   &
  \multicolumn{1}{c|}{} &
  \multicolumn{1}{c|}{3.81 (3.79)} &
  \multicolumn{1}{c|}{4.65 (3.95)} & 0.217
   \\ \cmidrule(r){1-5} \cmidrule(l){7-9} 
\multicolumn{2}{|c|}{CLAUSE (mean (SD))} &
  \multicolumn{1}{c|}{20.13 (9.22)} &
  \multicolumn{1}{c|}{20.43 (11.00)} & 0.030
   &
  \multicolumn{1}{c|}{} &
  \multicolumn{1}{c|}{19.72 (8.92)} &
  \multicolumn{1}{c|}{18.87 (9.21)} & 0.093
   \\ \cmidrule(r){1-5} \cmidrule(l){7-9} 
\multicolumn{2}{|c|}{AUX (mean (SD))} &
  \multicolumn{1}{c|}{13.18 (6.36)} &
  \multicolumn{1}{c|}{11.66 (7.09)} & 0.224
   &
  \multicolumn{1}{c|}{} &
  \multicolumn{1}{c|}{13.02 (6.32)} & 
  \multicolumn{1}{c|}{11.22 (6.55)} & 0.281
   \\ \cmidrule(r){1-5} \cmidrule(l){7-9} 
\multicolumn{2}{|c|}{VERB (mean (SD))} &
  \multicolumn{1}{c|}{16.81 (8.17)} &
  \multicolumn{1}{c|}{15.70 (8.79)} & 0.131
   &
  \multicolumn{1}{c|}{} &
  \multicolumn{1}{c|}{16.49 (7.94)} &
  \multicolumn{1}{c|}{15.00 (8.06)} & 0.186
   \\ \cmidrule(r){1-5} \cmidrule(l){7-9} 
\multicolumn{2}{|c|}{ADP (mean (SD))} &
  \multicolumn{1}{c|}{11.58 (7.22)} &
  \multicolumn{1}{c|}{9.29 (6.32)} & 0.338
   &
  \multicolumn{1}{c|}{} &
  \multicolumn{1}{c|}{11.35 (7.05)} &
  \multicolumn{1}{c|}{9.50 (6.40)} & 0.274
   \\ \cmidrule(r){1-5} \cmidrule(l){7-9} 
\multicolumn{2}{|c|}{DET (mean (SD))} &
  \multicolumn{1}{c|}{16.65 (9.07)} &
  \multicolumn{1}{c|}{13.73 (7.97)} & 0.342
   &
  \multicolumn{1}{c|}{} &
  \multicolumn{1}{c|}{16.40 (9.01)} &
  \multicolumn{1}{c|}{13.79 (8.05)} & 0.306
   \\ \cmidrule(r){1-5} \cmidrule(l){7-9} 
\multicolumn{2}{|c|}{PUNCT (mean (SD))} &
  \multicolumn{1}{c|}{24.41 (10.80)} &
  \multicolumn{1}{c|}{23.96 (12.13)} & 0.040
   &
  \multicolumn{1}{c|}{} &
  \multicolumn{1}{c|}{24.23 (10.67)} &
  \multicolumn{1}{c|}{22.23 (9.90)} & 0.195
   \\ \cmidrule(r){1-5} \cmidrule(l){7-9} 
\multicolumn{2}{|c|}{CCONJ (mean (SD))} &
  \multicolumn{1}{c|}{5.68 (4.28)} &
  \multicolumn{1}{c|}{5.84 (4.15)} & 0.038
   &
  \multicolumn{1}{c|}{} &
  \multicolumn{1}{c|}{5.59 (4.20)} &
  \multicolumn{1}{c|}{5.85 (4.21)} & 0.063
   \\ \cmidrule(r){1-5} \cmidrule(l){7-9} 
\multicolumn{2}{|c|}{PART (mean (SD))} &
  \multicolumn{1}{c|}{2.77 (2.25)} &
  \multicolumn{1}{c|}{3.21 (2.74)} & 0.174
   &
  \multicolumn{1}{c|}{} &
  \multicolumn{1}{c|}{2.59 (2.11)} &
  \multicolumn{1}{c|}{3.09 (2.50)} & 0.214
   \\ \cmidrule(r){1-5} \cmidrule(l){7-9} 
\multicolumn{2}{|c|}{SCONJ (mean (SD))} &
  \multicolumn{1}{c|}{1.63 (2.46)} &
  \multicolumn{1}{c|}{1.27 (1.78)} & 0.171
   &
  \multicolumn{1}{c|}{} &
  \multicolumn{1}{c|}{1.58 (2.46)} &
  \multicolumn{1}{c|}{1.18 (1.72)} & 0.189
   \\ \cmidrule(r){1-5} \cmidrule(l){7-9} 
\multicolumn{2}{|c|}{INTJ (mean (SD))} &
  \multicolumn{1}{c|}{5.16 (4.02)} &
  \multicolumn{1}{c|}{6.21 (6.83)} & 0.187
   &
  \multicolumn{1}{c|}{} &
  \multicolumn{1}{c|}{5.07 (3.97)} &
  \multicolumn{1}{c|}{5.66 (4.61)} & 0.138
   \\ \cmidrule(r){1-5} \cmidrule(l){7-9} 
\multicolumn{2}{|c|}{LF (mean (SD))} &
  \multicolumn{1}{c|}{8.16 (0.36)} &
  \multicolumn{1}{c|}{8.36 (0.47)} & 0.479
   &
  \multicolumn{1}{c|}{} &
  \multicolumn{1}{c|}{8.15 (0.37)} &
  \multicolumn{1}{c|}{8.30 (0.45)} & 0.358
   \\ \cmidrule(r){1-5} \cmidrule(l){7-9} 
\multicolumn{2}{|c|}{TTR (mean (SD))} &
  \multicolumn{1}{c|}{0.33 (0.05)} &
  \multicolumn{1}{c|}{0.31 (0.06)} & 0.373
   &
  \multicolumn{1}{c|}{} &
  \multicolumn{1}{c|}{0.34 (0.05)} &
  \multicolumn{1}{c|}{0.32 (0.06)} & 0.286
   \\ \cmidrule(r){1-5} \cmidrule(l){7-9} 
\multicolumn{2}{|c|}{par\_turns (mean (SD))} &
  \multicolumn{1}{c|}{13.55 (6.04)} &
  \multicolumn{1}{c|}{13.54 (6.98)} & 0.003
   &
  \multicolumn{1}{c|}{} &
  \multicolumn{1}{c|}{13.44 (5.97)} &
  \multicolumn{1}{c|}{12.38 (5.60)} & 0.183
   \\ \cmidrule(r){1-5} \cmidrule(l){7-9} 
\multicolumn{2}{|c|}{inv\_turns (mean (SD))} &
  \multicolumn{1}{c|}{3.16 (1.77)} &
  \multicolumn{1}{c|}{6.10 (4.48)} & 0.863
   &
  \multicolumn{1}{c|}{} &
  \multicolumn{1}{c|}{3.33 (1.73)} &
  \multicolumn{1}{c|}{4.38 (1.85)} & 0.589
   \\ \cmidrule(r){1-5} \cmidrule(l){7-9} 
\multicolumn{2}{|c|}{mmse (mean (SD))} &
  \multicolumn{1}{c|}{29.13 (1.11)} &
  \multicolumn{1}{c|}{18.54 (5.11)} & 2.864
   &
  \multicolumn{1}{c|}{} &
  \multicolumn{1}{c|}{29.08 (1.13)} &
  \multicolumn{1}{c|}{19.50 (4.50)} & 2.920
   \\ \bottomrule
\end{tabular}
}
\caption{The differences of linguistic features before/after matching on the Pitt corpus}
\label{tab:total_pitt}
\end{table*}

\begin{table*}[ht]
\centering
\small
\resizebox{0.7\linewidth}{!}{%
\begin{tabular}{@{}|ccclc|cccc|@{}}
\toprule
\multicolumn{1}{|c|}{\multirow{2}{*}{\textbf{Features}}} &
  \multicolumn{4}{c|}{\textbf{\textit{Before matching}}} &
  \multicolumn{4}{c|}{\textbf{\textit{After matching}}} \\ \cmidrule(l){2-9} 
\multicolumn{1}{|c|}{} &
  \multicolumn{1}{c|}{Level} &
  \multicolumn{1}{c|}{Control} &
  \multicolumn{1}{c|}{Dementia} &
 SMD  &
  \multicolumn{1}{c|}{\multirow{2}{*}{Level}} &
  \multicolumn{1}{c|}{Control} &
  \multicolumn{1}{c|}{Dementia} & SMD
   \\ \cmidrule(r){1-5} \cmidrule(l){7-9} 
\multicolumn{2}{|c|}{Number of transcripts ($n$)} &
  \multicolumn{1}{c|}{1017} &
  \multicolumn{1}{c|}{152} &
   &
  \multicolumn{1}{c|}{} &
  \multicolumn{1}{c|}{152} &
  \multicolumn{1}{c|}{152} &
   \\ \midrule
\multicolumn{2}{|c|}{Education (mean (SD))} &
  \multicolumn{1}{c|}{13.77 (3.01)} &
  \multicolumn{1}{c|}{12.64 (2.16)} & 0.431
   &
  \multicolumn{1}{c|}{} &
  \multicolumn{1}{c|}{12.62 (2.18)} &
  \multicolumn{1}{c|}{12.64 (2.16)} & 0.006
   \\ \cmidrule(r){1-5} \cmidrule(l){7-9} 
\multicolumn{2}{|c|}{Age (mean (SD))} &
  \multicolumn{1}{c|}{70.30 (4.14)} &
  \multicolumn{1}{c|}{70.20 (5.75)} & 0.021
   & 
  \multicolumn{1}{c|}{} &
  \multicolumn{1}{c|}{70.81 (3.77)} &
  \multicolumn{1}{c|}{70.20 (5.75)} & 0.126
   \\ \cmidrule(r){1-5} \cmidrule(l){7-9} 
\multicolumn{2}{|c|}{PRON (mean (SD))} &
  \multicolumn{1}{c|}{15.20 (9.93)} &
  \multicolumn{1}{c|}{11.16 (8.05)} & 0.447
   &
  \multicolumn{1}{c|}{\multirow{20}{*}{}} &
  \multicolumn{1}{c|}{14.37 (8.91)} &
  \multicolumn{1}{c|}{11.16 (8.05)} & 0.377
   \\ \cmidrule(r){1-5} \cmidrule(l){7-9} 
\multicolumn{2}{|c|}{AUX (mean (SD))} &
  \multicolumn{1}{c|}{10.76 (6.63)} &
  \multicolumn{1}{c|}{7.81 (5.48)} & 0.485
   &
  \multicolumn{1}{c|}{} &
  \multicolumn{1}{c|}{9.75 (6.25)} &
  \multicolumn{1}{c|}{7.81 (5.48)} & 0.330
   \\ \cmidrule(r){1-5} \cmidrule(l){7-9} 
\multicolumn{2}{|c|}{VERB (mean (SD))} &
  \multicolumn{1}{c|}{16.82 (9.08)} &
  \multicolumn{1}{c|}{12.71 (7.19)} & 0.502
   &
  \multicolumn{1}{c|}{} &
  \multicolumn{1}{c|}{15.25 (7.91)} &
  \multicolumn{1}{c|}{12.71 (7.19)} & 0.336
   \\ \cmidrule(r){1-5} \cmidrule(l){7-9} 
\multicolumn{2}{|c|}{ADP (mean (SD))} &
  \multicolumn{1}{c|}{11.53 (6.74)} &
  \multicolumn{1}{c|}{8.57 (5.71)} & 0.474
   &
  \multicolumn{1}{c|}{} &
  \multicolumn{1}{c|}{10.07 (5.77)} &
  \multicolumn{1}{c|}{8.57 (5.71)} & 0.262
   \\ \cmidrule(r){1-5} \cmidrule(l){7-9} 
\multicolumn{2}{|c|}{DET (mean (SD))} &
  \multicolumn{1}{c|}{16.99 (9.87)} &
  \multicolumn{1}{c|}{12.22 (7.23)} & 0.551
   &
  \multicolumn{1}{c|}{} &
  \multicolumn{1}{c|}{15.03 (8.00)} &
  \multicolumn{1}{c|}{12.22 (7.23)} & 0.368
   \\ \cmidrule(r){1-5} \cmidrule(l){7-9} 
\multicolumn{2}{|c|}{NOUN (mean (SD))} &
  \multicolumn{1}{c|}{29.00 (16.89)} &
  \multicolumn{1}{c|}{22.38 (13.71)} & 0.430
   &
  \multicolumn{1}{c|}{} &
  \multicolumn{1}{c|}{26.86 (14.61)} &
  \multicolumn{1}{c|}{22.38 (13.71)} & 0.316
   \\ \cmidrule(r){1-5} \cmidrule(l){7-9} 
\multicolumn{2}{|c|}{PUNCT (mean (SD))} &
  \multicolumn{1}{c|}{26.61 (13.24)} &
  \multicolumn{1}{c|}{21.68 (11.74)} & 0.393
   &
  \multicolumn{1}{c|}{} &
  \multicolumn{1}{c|}{25.47 (11.87)} &
  \multicolumn{1}{c|}{21.68 (11.74)} & 0.321
   \\ \cmidrule(r){1-5} \cmidrule(l){7-9} 
\multicolumn{2}{|c|}{CCONJ (mean (SD))} &
  \multicolumn{1}{c|}{5.47 (4.83)} &
  \multicolumn{1}{c|}{3.43 (3.62)} & 0.478
   &
  \multicolumn{1}{c|}{} &
  \multicolumn{1}{c|}{5.02 (4.99)} &
  \multicolumn{1}{c|}{3.43 (3.62)} & 0.365
   \\ \cmidrule(r){1-5} \cmidrule(l){7-9} 
\multicolumn{2}{|c|}{ADJ (mean (SD))} &
  \multicolumn{1}{c|}{3.89 (3.75)} &
  \multicolumn{1}{c|}{2.30 (2.37)} & 0.509
   &
  \multicolumn{1}{c|}{} &
  \multicolumn{1}{c|}{3.02 (2.93)} &
  \multicolumn{1}{c|}{2.30 (2.37)} & 0.272
   \\ \cmidrule(r){1-5} \cmidrule(l){7-9} 
\multicolumn{2}{|c|}{PART (mean (SD))} &
  \multicolumn{1}{c|}{2.83 (2.46)} &
  \multicolumn{1}{c|}{2.40 (2.14)} & 0.186
   &
  \multicolumn{1}{c|}{} &
  \multicolumn{1}{c|}{2.63 (2.25)} &
  \multicolumn{1}{c|}{2.40 (2.14)} & 0.105
   \\ \cmidrule(r){1-5} \cmidrule(l){7-9} 
\multicolumn{2}{|c|}{SCONJ (mean (SD))} &
  \multicolumn{1}{c|}{1.55 (1.92)} &
  \multicolumn{1}{c|}{0.90 (1.36)} & 0.390
   &
  \multicolumn{1}{c|}{} &
  \multicolumn{1}{c|}{1.38 (1.60)} &
  \multicolumn{1}{c|}{0.90 (1.36)} & 0.324
   \\ \cmidrule(r){1-5} \cmidrule(l){7-9} 
\multicolumn{2}{|c|}{ADV (mean (SD))} &
  \multicolumn{1}{c|}{4.06 (3.98)} &
  \multicolumn{1}{c|}{2.93 (3.41)} & 0.305
   &
  \multicolumn{1}{c|}{} &
  \multicolumn{1}{c|}{3.77 (3.60)} &
  \multicolumn{1}{c|}{2.93 (3.41)} & 0.240
   \\ \cmidrule(r){1-5} \cmidrule(l){7-9} 
\multicolumn{2}{|c|}{INTJ (mean (SD))} &
  \multicolumn{1}{c|}{1.88 (2.91)} &
  \multicolumn{1}{c|}{1.50 (2.58)} & 0.139
   &
  \multicolumn{1}{c|}{} &
  \multicolumn{1}{c|}{1.99 (2.80)} &
  \multicolumn{1}{c|}{1.50 (2.58)} & 0.181
   \\ \cmidrule(r){1-5} \cmidrule(l){7-9} 
\multicolumn{2}{|c|}{LF (mean (SD))} &
  \multicolumn{1}{c|}{8.06 (0.43)} &
  \multicolumn{1}{c|}{8.05 (0.44)} & 0.033
   &
  \multicolumn{1}{c|}{} &
  \multicolumn{1}{c|}{8.02 (0.42)} &
  \multicolumn{1}{c|}{8.05 (0.44)} & 0.057
   \\ \cmidrule(r){1-5} \cmidrule(l){7-9} 
\multicolumn{2}{|c|}{TTR (mean (SD))} &
  \multicolumn{1}{c|}{0.35 (0.06)} &
  \multicolumn{1}{c|}{0.37 (0.07)} & 0.303
   &
  \multicolumn{1}{c|}{} &
  \multicolumn{1}{c|}{0.36 (0.06)} &
  \multicolumn{1}{c|}{0.37 (0.07)} & 0.189
   \\ \cmidrule(r){1-5} \cmidrule(l){7-9} 
\multicolumn{2}{|c|}{CLAUSE (mean (SD))} &
  \multicolumn{1}{c|}{21.01 (10.44)} &
  \multicolumn{1}{c|}{17.21 (9.26)} & 0.385
   &
  \multicolumn{1}{c|}{} &
  \multicolumn{1}{c|}{19.81 (9.14)} &
  \multicolumn{1}{c|}{17.21 (9.26)} & 0.282
   \\ \cmidrule(r){1-5} \cmidrule(l){7-9} 
\multicolumn{2}{|c|}{PROPN (mean (SD))} &
  \multicolumn{1}{c|}{0.07 (0.36)} &
  \multicolumn{1}{c|}{0.02 (0.18)} & 0.170
   &
  \multicolumn{1}{c|}{} &
  \multicolumn{1}{c|}{0.14 (0.55)} &
  \multicolumn{1}{c|}{0.02 (0.18)} & 0.288
   \\ \cmidrule(r){1-5} \cmidrule(l){7-9} 
\multicolumn{2}{|c|}{par\_turns (mean (SD))} &
  \multicolumn{1}{c|}{14.39 (7.91)} &
  \multicolumn{1}{c|}{11.97 (7.04)} & 0.323
   &
  \multicolumn{1}{c|}{} &
  \multicolumn{1}{c|}{13.88 (6.68)} &
  \multicolumn{1}{c|}{11.97 (7.04)} & 0.278
   \\ \cmidrule(r){1-5} \cmidrule(l){7-9} 
\multicolumn{2}{|c|}{inv\_turns (mean (SD))} &
  \multicolumn{1}{c|}{0.75 (1.53)} &
  \multicolumn{1}{c|}{0.82 (1.79)} & 0.044
   &
  \multicolumn{1}{c|}{} &
  \multicolumn{1}{c|}{0.77 (1.25)} &
  \multicolumn{1}{c|}{0.82 (1.79)} & 0.034
   \\ \bottomrule
\end{tabular}
}
\caption{The differences of linguistic features before/after matching on the WLS dataset}
\label{tab:total_wls}
\end{table*}

\end{document}